\pdfoutput=1
\PassOptionsToPackage{numbers,square,sort&compress}{natbib}
\PassOptionsToPackage{table,dvipsnames}{xcolor}
\documentclass[]{style/company_light} %

\usepackage[utf8]{inputenc}
\usepackage{hyperref}
\usepackage{url}
\usepackage{booktabs}
\usepackage{amsfonts}
\usepackage{amsmath}
\usepackage{nicefrac}
\usepackage{microtype}
\usepackage{graphicx}
\usepackage{float}
\usepackage{subcaption}
\usepackage{multirow}
\usepackage{enumerate}
\usepackage{enumitem}
\usepackage[most]{tcolorbox}
\usepackage{adjustbox}
\usepackage{array}
\usepackage{amssymb}
\usepackage{tabularx}
\usepackage{longtable}
\usepackage{listings}
\usepackage{needspace}

\makeatletter
\def\fps@figure{htbp}
\def\fps@table{htbp}
\makeatother

\newcommand{\compactfloatlayout}{%
  \renewcommand{\topfraction}{0.92}%
  \renewcommand{\bottomfraction}{0.88}%
  \renewcommand{\textfraction}{0.08}%
  \renewcommand{\floatpagefraction}{0.85}%
  \setcounter{topnumber}{3}%
  \setcounter{bottomnumber}{2}%
  \setcounter{totalnumber}{4}%
  \setlength{\textfloatsep}{10pt plus 2pt minus 2pt}%
  \setlength{\intextsep}{10pt plus 2pt minus 2pt}%
  \setlength{\floatsep}{8pt plus 2pt minus 2pt}%
  \interlinepenalty=0\relax
  \clubpenalty=10000\relax
  \widowpenalty=10000\relax
  \raggedbottom
}

\NewDocumentCommand{\todo}
{ mO{} }{\textcolor{magenta}{\textsuperscript{\textit{TODO}}\textsf{\textbf{\small[#1]}}}}

\title{Visual Cipher Bench: Benchmarking LLM Agents on Long-Horizon Visual Reasoning}
\author[*,\dagger]{Deyao Hong}
\author[*]{Jiayi Jiang}
\author[*]{Yuchun Chen}
\author{Yixuan Shi}
\author{Wenyi Li}
\author{Zheng Jiang}
\author{Chengxing Xie}
\author{Mingyu Su}
\author{Qingle Liu}
\author{Mingju Gao}
\author{Yizhe Chi}
\author{Youjie Zheng}
\author[\ddagger]{Calvin Xiao}
\author[\ddagger]{Kaisen Yang}
\author[\ddagger]{Huan-ang Gao}

\affiliation{Navers Lab, Einsia.AI,\quad\quad Tsinghua University}
\contribution[*]{Equal Contribution}
\contribution[\dagger]{Project Lead}
\contribution[\ddagger]{Corresponding Authors}

\abstract{%
While language-model agents excel at long-horizon coding and terminal tasks, this success relies on two conditions: explicit text specifications and immediate execution feedback. However, neither condition holds in real-world problems where critical information exists only in images, leaving it unclear whether long-horizon capability transfers from text to visual reasoning.
In such visual settings, because rules and clues resist text transcription and lack test-case feedback, agents must repeatedly inspect images, simulate spatial transformations, and autonomously cross-check clues.
Despite this challenge, existing evaluations fail to capture such capability: visual reasoning benchmarks largely reduce tasks to single-turn, short-chain question answering, whereas agent benchmarks focus on text-based environments or interactive states that permit trial-and-error.
To bridge this gap, we introduce \textbf{Visual Cipher Bench}, a collection of 40 original visual puzzles across four categories: spatial manipulation and simulation, logical constraints and planning, rule induction, and knowledge-based decoding. Each puzzle provides only one to three black-and-white line drawings and a minimal submission instruction, requiring agents to discover the underlying solving mechanism entirely from the images.
Across 29 reasoning-effort configurations of eight frontier models, even the strongest model, GPT-6 Astra (max), achieves only 26.7\% accuracy, while five of the eight models score below 1\%. Yet on a ten-task subset, providing a single non-spoiler mechanism hint boosts accuracy from 12.2\% to 51.1\%.
This sharp contrast shows that the primary bottleneck of current agents lies not in executing multi-step plans, but in autonomously discovering rules from visual evidence.
Visual Cipher Bench thus provides a rigorous benchmark for measuring long-horizon visual reasoning in agents.
}

\date{\today}
\metadata[Homepage]{\url{https://lab.einsia.ai/visual-cipher-bench}}
\correspondence{\email{nana@einsia.ai}}

\begin{document}

\maketitle

\compactfloatlayout
\section{Introduction}
\label{sec:introduction}

\begin{figure*}[t]
  \centering
  \includegraphics[width=\linewidth]{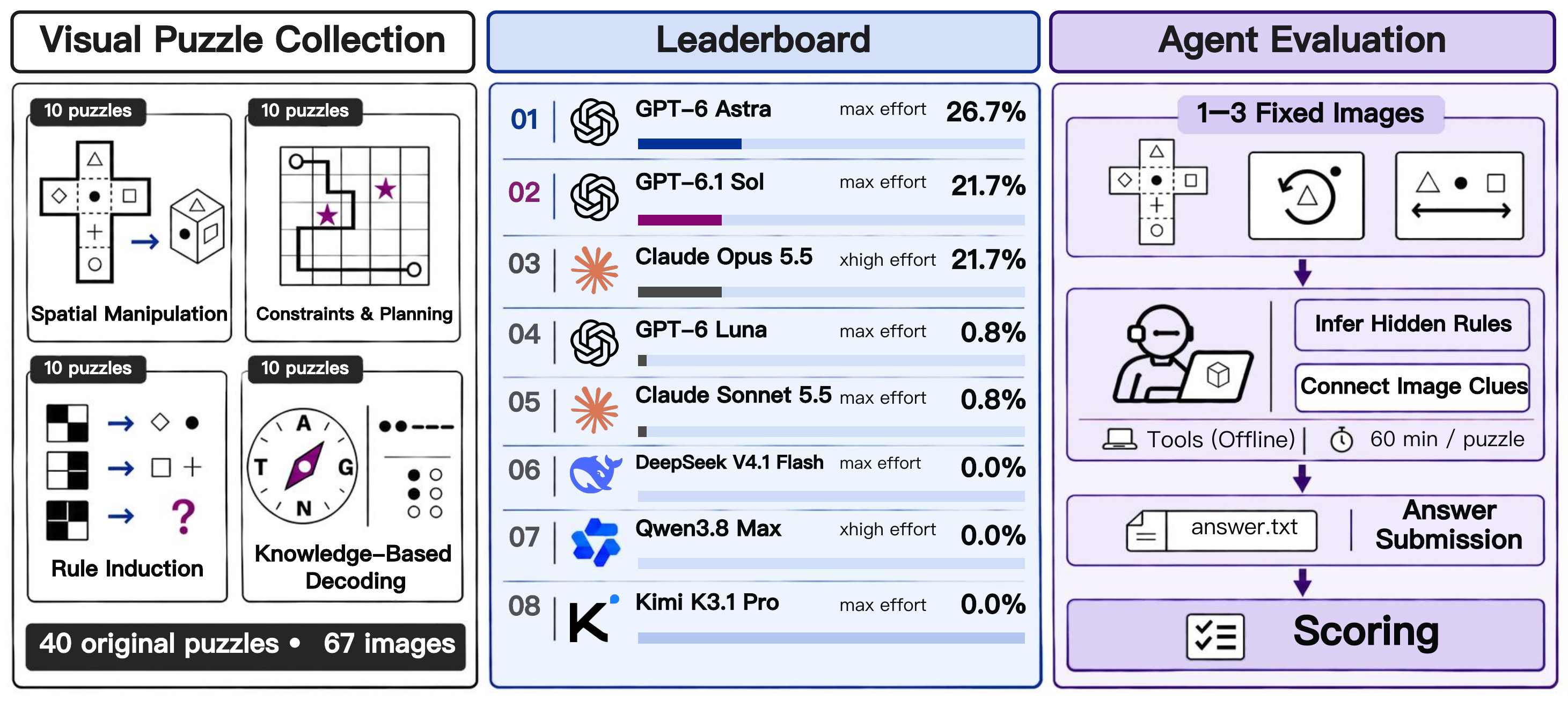}
  \caption{Overview of Visual Cipher Bench. \textbf{Left:} 40 original visual puzzles (67 images in total), divided by primary difficulty into spatial manipulation and simulation, logical constraints and planning, rule induction, and knowledge-based decoding, with 10 puzzles each. \textbf{Middle:} mean accuracy of each model's best reasoning-effort configuration over three runs. \textbf{Right:} the evaluation pipeline.}
  \label{fig:framework}
\end{figure*}

Language-model agents have achieved remarkable long-horizon reasoning in coding and terminal tasks: on benchmarks such as SWE-Bench Pro and Terminal-Bench 4.0, they autonomously resolve complex issues by iteratively running tests, reading error logs, and revising their plans ~\citep{deng2025swebenchpro,marten2026terminalbench40,hong2026swerefactorbench,chan2025mlebench,chi2026frontiereng,chi2026ai4aibench,kwa2025measuring}. This success, however, rests on two critical conditions: task specifications are explicitly provided as readable text or code, and execution results immediately signal whether the current trajectory is correct. In many real-world problems, critical information exists only in images. Although frontier models possess foundational visual perception~\citep{openai2023gpt4,comanici2025gemini,bai2025qwen3vl} and can invoke tools to crop, annotate, and transform images during reasoning~\citep{gupta2023visprog,hu2024visualsketchpad,zheng2025deepeyes}, it remains unclear whether long-horizon reasoning capability truly transfers from text to the visual domain.

The visual setting is uniquely challenging because neither condition holds. First, rules, examples, and clues are embedded in static images that resist direct text transcription, while multimodal models still falter on basic perceptual judgments~\citep{fu2024blink,tong2024eyes,rahmanzadehgervi2024blind}. Second, the environment provides no test-case feedback during problem solving, forcing agents to validate hypotheses solely by cross-checking internal visual clues. Consider folding a symbol-marked cube net: even after correctly computing face adjacencies, the solver must still determine the starting face and reading order by re-examining the image for subtle cues. Without external verification, an agent can easily halt after a locally valid step or prematurely commit to a flawed rule simply because a few decoded letters resemble a word. Consequently, long-horizon visual reasoning cannot be reduced to a single pass; it requires agents to iteratively cycle between visual inspection and hypothesis testing---proposing rules, verifying them against global visual constraints, and discarding false assumptions until converging on a unique solution.

Existing evaluations fail to capture this capability. Standard visual reasoning benchmarks reduce problem solving to single-turn question answering with explicitly defined tasks~\citep{yue2024mmmu,lu2024mathvista,yue2025mmmupro,zhang2024mathverse,li2026benchmarking}. Even when benchmarks incorporate abstract reasoning or spatial simulation, their fixed task formats already telegraph where the underlying rule lies~\citep{zhang2019raven,chollet2025arcagi2,li2026stare}. Meanwhile, agent benchmarks focus predominantly on text and code environments~\citep{zhou2024webarena,mialon2024gaia,jimenez2024swebench}. When these benchmarks do involve vision, images either serve merely as targets for GUI localization or appear in interactive environments that return a new state after every action, allowing agents to rely on trial-and-error rather than pure visual deduction ~\citep{xie2024osworld,paglieri2025balrog,zhang2026longpuzzlebench,arcprize2026arcagi3}. Consequently, the field lacks a dedicated benchmark for evaluating long-horizon reasoning over static visual evidence.

To bridge this gap, we introduce \textbf{Visual Cipher Bench} (Figure~\ref{fig:framework}). Visual puzzles are an ideal testbed for long-horizon visual reasoning: with all rules and clues concealed in the graphics, solvers must uncover the mechanism themselves, as each discovery dictates what to inspect and deduce next~\citep{cj2020extraction,puzzledpintauthoring}.
The benchmark comprises 40 original, expert-verified puzzles across four categories: spatial manipulation and simulation, logical constraints and planning, rule induction, and knowledge-based decoding. Each puzzle provides only one to three black-and-white line drawings and a minimal submission instruction. Agents have 60 minutes in an offline coding sandbox to solve each puzzle and are graded strictly by exact match on the final answer, while hidden walkthroughs and reference solvers ensure solvability and enable process auditing.

We systematically evaluate 29 reasoning-effort configurations across eight frontier models~\citep{anthropic2026opus55,anthropic2026sonnet55,openai2026gpt61sol,openai2026gpt6astra,openai2026gpt6luna,deepseek2026v41flash,qwen2026qwen38max,kimi2026k31pro}, conducting three independent runs per configuration. Even the strongest configuration, GPT-6 Astra (max), achieves only 26.7\% accuracy, while five of the eight models score below 1\% in their best configurations. Moreover, performance is heavily skewed across categories: pooled accuracy reaches 22.8\% on spatial manipulation and simulation, where transformation rules are visually explicit, but drops to just 0.6\% on rule induction. Yet on a ten-task subset spanning all four categories, providing a single non-spoiler mechanism hint boosts accuracy from 12.2\% to 51.1\% and cuts timeouts from 51 to 29 under the same time limit. This sharp contrast demonstrates that the primary bottleneck of current agents lies not in executing multi-step plans, but in autonomously inducing rules from visual evidence.

Our contributions are threefold:
\begin{itemize}
    \item We introduce Visual Cipher Bench, a long-horizon visual reasoning benchmark of 40 original puzzles that embeds all rules and clues exclusively within static images and provides no intermediate feedback during solving.
    \item We benchmark 29 reasoning-effort configurations across eight frontier models, exposing a stark capability ceiling: even the top-performing agent solves barely a quarter of the tasks, while most frontier models fail almost completely.
    \item We pinpoint the core bottleneck of visual agents through hint experiments and trajectory audits, demonstrating that agents fail primarily at inducing rules and constraints from visual evidence rather than executing multi-step plans.
\end{itemize}

\section{Visual Cipher Bench}
\label{sec:vcb-benchmark}

\subsection{Task Formulation and Evaluation}
\label{sec:vcb-task}
\label{sec:vcb-evaluation}

Each task is denoted \(\tau=(I,q,a^\star)\), where \(I=(I_1,\ldots,I_k)\) contains \(k\in\{1,2,3\}\) puzzle images, \(q\) is the submission instruction shared by all tasks (Appendix~\ref{app:eval-setup}), and \(a^\star\) is the unique reference answer. The agent observes only \(I\) and \(q\). The instruction only asks the agent to inspect the images and submit a single final answer, without hinting at any solution method: the agent must infer from the images which mechanism to use, which graphics are examples and which are inputs to process, and whether the answer should be a word or a symbol sequence.

Solving is a long-horizon interactive process. The agent works in a network-disabled sandbox; at each step it may inspect the images, write and run code, and decide what to do next based on the execution results, submitting a final answer \(\hat a\) within a fixed time limit. No correctness feedback is provided during solving, so hypotheses can be checked only against other clues in the images. The runtime harness, time limit, and compute resources are listed in Appendix~\ref{app:eval-setup}.

Grading considers only the final answer. Let \(N(\cdot)\) be an answer-normalization function fixed before evaluation (e.g., removing whitespace and unifying letter case). The score of a single attempt is
\begin{equation}
  s_{\tau}=\mathbf{1}\!\left[N(\hat a)=N(a^\star)\right],
  \label{eq:vcb-score}
\end{equation}
and timeouts and invalid submissions also score 0. Grading is deterministic and runs in a separate container, so the agent cannot read or modify the reference answer~\citep{wang2026benchjack}, and no LLM judge is involved~\citep{zheng2023judging}. Each model--effort configuration is run independently three times on all 40 tasks. Letting \(s_{ir}\in\{0,1\}\) denote the score on task \(i\) in run \(r\), we report mean accuracy
\begin{equation}
  \mathrm{Acc}=\frac{1}{120}\sum_{i=1}^{40}\sum_{r=1}^{3}s_{ir};
  \label{eq:vcb-metrics}
\end{equation}
category accuracy is computed in the same way over the \(10\times3=30\) attempts in each category.

\subsection{Benchmark Design and Composition}
\label{sec:vcb-construction}

\paragraph{Design principles.}
We surveyed hundreds of public puzzle resources, including past puzzle-competition problems, solver write-ups, and authoring guides~\citep{cj2020extraction,puzzledpintauthoring}, and designed the puzzles according to two principles. The first is solvability: every key inference is supported by evidence on the puzzle page, the clues suffice to determine a unique answer, and solvers need neither guesswork nor hints from outside the puzzle. The second is long-horizon reasoning: the answer cannot be read off in a single step; solvers must first discover the mechanism, then work through multiple interdependent steps and check earlier inferences against later clues, and the answer cannot be guessed from local clues or thematic association alone. The former ensures puzzle quality; the latter ensures difficulty.

\paragraph{Originality and review.}
All 40 puzzles are original to this benchmark. Each puzzle was designed together with a complete solution, consisting of a step-by-step walkthrough and a reference solver: the former records the visual evidence behind each key step, and the latter reproduces the reference answer using only the puzzle images. Every puzzle was reviewed by human experts, who confirmed that the solution path is sound, the visual evidence is sufficient, and the answer is unique. Neither material is provided to evaluated agents.

\paragraph{Puzzle categories.}
We categorize puzzles along two dimensions: whether the main difficulty lies in executing an identified mechanism or in discovering which mechanism to use, and whether solving relies mainly on rule-constrained reasoning or on visual imagination and knowledge association. Combining the two dimensions yields four primary categories; each puzzle belongs to one category, and each category contains 10 puzzles (Figure~\ref{fig:framework}):
\begin{itemize}[leftmargin=*,nosep]
  \item P1 Spatial manipulation and simulation: simulate the state after operations such as folding, rotation, and projection;
  \item P2 Logical constraints and planning: carry out derivations under grid, path, or state constraints;
  \item P3 Rule induction: induce a puzzle-specific rule or cipher from examples within the puzzle;
  \item P4 Knowledge-based decoding: recognize the external convention indicated by the visual clues, such as Morse code, Braille, or wordplay, and decode accordingly.
\end{itemize}

\paragraph{Dataset composition.}
The 40 tasks contain 67 images, all drawn as black-and-white line art in a consistent style: 19 single-image tasks, 15 two-image tasks, and six three-image tasks. Reference answers range from 4 to 30 characters. Twenty-eight are English words or phrases; the other 12 include digit strings, letter codes, and related sequences for which the exact characters and their order must be preserved.

\definecolor{vcbPerformance}{HTML}{E8AF7A}
\definecolor{vcbCorrect}{HTML}{8FC29F}
\definecolor{vcbWrong}{HTML}{D9958E}
\definecolor{vcbTimeout}{HTML}{E3C66F}
\definecolor{vcbError}{HTML}{B8C0C3}
\definecolor{vcbBestRow}{HTML}{F4E7D7}
\definecolor{vcbHeader}{HTML}{EEEAE4}
\definecolor{vcbSpatial}{HTML}{DDF1EC}
\definecolor{vcbLogic}{HTML}{E4ECF8}
\definecolor{vcbRule}{HTML}{EEE5F7}
\definecolor{vcbKnowledge}{HTML}{F8E9D8}

\definecolor{vcbScoreCol}{HTML}{F1F1F1}
\definecolor{vcbBestRowLight}{HTML}{F1F1F1}
\newcommand{\vcblogo}[1]{\raisebox{-0.2em}{\includegraphics[height=1.05em]{Figures/logos/#1}}\hspace{0.45em}}

\newcommand{\vcbperf}[2]{\cellcolor{vcbPerformance!#1}\strut #2}
\newcommand{\vcbcorrect}[2]{\cellcolor{vcbCorrect!#1}\strut #2}
\newcommand{\vcbwrong}[2]{\cellcolor{vcbWrong!#1}\strut #2}
\newcommand{\vcbtimeout}[2]{\cellcolor{vcbTimeout!#1}\strut #2}
\newcommand{\vcberror}[2]{\cellcolor{vcbError!#1}\strut #2}
\newcommand{\vcbPending}{\textsuperscript{\textcolor{red}{\tiny TODO}}}
\newcommand{\vcbtile}[1]{{\setlength{\fboxsep}{0pt}\colorbox{#1}{\rule{1.55ex}{0pt}\rule{0pt}{1.55ex}}}}
\newcommand{\vcbtriplet}[3]{\vcbtile{#1}\hspace{0.35ex}\vcbtile{#2}\hspace{0.35ex}\vcbtile{#3}}
\newcommand{\vcbcategory}[2]{{\setlength{\fboxsep}{2pt}\colorbox{#1}{\strut\scriptsize #2}}}
\newcommand{\vcbscorecell}[1]{%
  \ifcase#1\cellcolor{vcbHeader!75}\textbf{0}%
  \or\cellcolor{vcbCorrect!35}\textbf{1}%
  \or\cellcolor{vcbCorrect!62}\textbf{2}%
  \or\cellcolor{vcbCorrect!88}\textbf{3}%
  \fi}

\usetikzlibrary{arrows.meta,calc}

\section{Experiments}

\subsection{Experimental Setup}

We evaluate eight models: Claude Opus 5.5~\citep{anthropic2026opus55}, Claude Sonnet 5.5~\citep{anthropic2026sonnet55}, GPT-6.1 Sol~\citep{openai2026gpt61sol}, GPT-6 Astra~\citep{openai2026gpt6astra}, GPT-6 Luna~\citep{openai2026gpt6luna}, DeepSeek V4.1 Flash~\citep{deepseek2026v41flash}, Qwen 3.8 Max~\citep{qwen2026qwen38max}, and Kimi K3.1 Pro~\citep{kimi2026k31pro}. Depending on the interfaces available for each model, we evaluate one or more reasoning-effort levels, yielding 29 model–effort configurations. Every configuration is run independently three times on all 40 tasks, for 87 complete benchmark runs and 3,480 task-level attempts.

\medskip
The evaluation protocol follows Section~\ref{sec:vcb-evaluation}; the runtime harness, time limit, and compute resources are listed in Appendix~\ref{app:eval-setup}. We report two metrics:

\medskip
\begin{enumerate}
  \item Mean accuracy: accuracy over the 120 attempts of a configuration across three runs (Eq.~\eqref{eq:vcb-metrics}); Table~\ref{tab:exp-32-t01} also reports the corresponding mean number of tasks solved per run (Mean /40).
  \item Category accuracy: computed in the same way for each of the four categories (spatial manipulation and simulation, logical constraints and planning, rule induction, and knowledge-based decoding) over the 30 attempts per category.
\end{enumerate}

\medskip
\subsection{Overall Performance}

\begin{table*}[t]
\centering
\caption{Results of all 29 configurations on Visual Cipher Bench, with models ordered by their best configuration. P1--P4 give the category accuracy over the 30 attempts in each category, Acc.\ gives the mean accuracy over all 120 attempts, and Mean gives the corresponding number of tasks solved per run (out of 40). All values average three runs; the best value in each column is in bold. Each model's best configuration is shaded.}
\label{tab:exp-32-t01}
\begingroup
\small
\setlength{\tabcolsep}{4pt}
\setlength{\aboverulesep}{0pt}
\setlength{\belowrulesep}{0pt}
\renewcommand{\arraystretch}{1.32}
\begin{tabularx}{\linewidth}{ll*{5}{>{\centering\arraybackslash}X}>{\centering\arraybackslash}X}
\toprule
& & \multicolumn{4}{c}{Category accuracy (\%) $\uparrow$} & \multicolumn{2}{c}{Overall $\uparrow$} \\
\cmidrule(lr){3-6}\cmidrule(l){7-8}
\rule{0pt}{4.4ex}Model & Effort & \shortstack{P1\\Spatial} & \shortstack{P2\\Logic} & \shortstack{P3\\Rule} & \shortstack{P4\\Knowledge} & Mean /40 & Acc.\ (\%) \\
\midrule
\rowcolor{vcbBestRowLight}\vcblogo{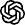}GPT-6 Astra & max & \textbf{80.0} & \textbf{13.3} & 0.0 & \textbf{13.3} & \textbf{10.7} & \textbf{26.7} \\
\vcblogo{openai}GPT-6 Astra & xhigh & 66.7 & 6.7 & \textbf{3.3} & \textbf{13.3} & 9.0 & 22.5 \\
\vcblogo{openai}GPT-6 Astra & high & 63.3 & 10.0 & \textbf{3.3} & 10.0 & 8.7 & 21.7 \\
\vcblogo{openai}GPT-6 Astra & medium & 30.0 & 6.7 & \textbf{3.3} & 6.7 & 4.7 & 11.7 \\
\vcblogo{openai}GPT-6 Astra & low & 20.0 & 0.0 & 0.0 & 0.0 & 2.0 & 5.0 \\
\arrayrulecolor{black!22}\midrule[0.3pt]\arrayrulecolor{black}
\rowcolor{vcbBestRowLight}\vcblogo{openai}GPT-6.1 Sol & max & 60.0 & 10.0 & \textbf{3.3} & \textbf{13.3} & 8.7 & 21.7 \\
\vcblogo{openai}GPT-6.1 Sol & xhigh & 53.3 & 6.7 & 0.0 & 3.3 & 6.3 & 15.8 \\
\vcblogo{openai}GPT-6.1 Sol & high & 36.7 & 3.3 & 0.0 & 6.7 & 4.7 & 11.7 \\
\vcblogo{openai}GPT-6.1 Sol & medium & 26.7 & 0.0 & 0.0 & 6.7 & 3.3 & 8.3 \\
\vcblogo{openai}GPT-6.1 Sol & low & 16.7 & 0.0 & 0.0 & 3.3 & 2.0 & 5.0 \\
\arrayrulecolor{black!22}\midrule[0.3pt]\arrayrulecolor{black}
\vcblogo{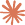}Claude Opus 5.5 & max & 56.7 & 6.7 & 0.0 & 6.7 & 7.0 & 17.5 \\
\rowcolor{vcbBestRowLight}\vcblogo{claude}Claude Opus 5.5 & xhigh & 70.0 & 6.7 & 0.0 & 10.0 & 8.7 & 21.7 \\
\vcblogo{claude}Claude Opus 5.5 & high & 40.0 & 0.0 & 0.0 & 3.3 & 4.3 & 10.8 \\
\vcblogo{claude}Claude Opus 5.5 & medium & 33.3 & 0.0 & 0.0 & 0.0 & 3.3 & 8.3 \\
\vcblogo{claude}Claude Opus 5.5 & low & 0.0 & 0.0 & 0.0 & 0.0 & 0.0 & 0.0 \\
\arrayrulecolor{black!22}\midrule[0.3pt]\arrayrulecolor{black}
\rowcolor{vcbBestRowLight}\vcblogo{openai}GPT-6 Luna & max & 3.3 & 0.0 & 0.0 & 0.0 & 0.3 & 0.8 \\
\vcblogo{openai}GPT-6 Luna & xhigh & 0.0 & 0.0 & 0.0 & 0.0 & 0.0 & 0.0 \\
\vcblogo{openai}GPT-6 Luna & high & 0.0 & 0.0 & 0.0 & 0.0 & 0.0 & 0.0 \\
\vcblogo{openai}GPT-6 Luna & medium & 0.0 & 0.0 & 0.0 & 0.0 & 0.0 & 0.0 \\
\vcblogo{openai}GPT-6 Luna & low & 0.0 & 0.0 & 0.0 & 0.0 & 0.0 & 0.0 \\
\vcblogo{openai}GPT-6 Luna & none & 0.0 & 0.0 & 0.0 & 0.0 & 0.0 & 0.0 \\
\arrayrulecolor{black!22}\midrule[0.3pt]\arrayrulecolor{black}
\rowcolor{vcbBestRowLight}\vcblogo{claude}Claude Sonnet 5.5 & max & 3.3 & 0.0 & 0.0 & 0.0 & 0.3 & 0.8 \\
\vcblogo{claude}Claude Sonnet 5.5 & xhigh & 0.0 & 0.0 & 0.0 & 0.0 & 0.0 & 0.0 \\
\vcblogo{claude}Claude Sonnet 5.5 & high & 0.0 & 0.0 & 0.0 & 0.0 & 0.0 & 0.0 \\
\vcblogo{claude}Claude Sonnet 5.5 & medium & 0.0 & 0.0 & 0.0 & 0.0 & 0.0 & 0.0 \\
\vcblogo{claude}Claude Sonnet 5.5 & low & 0.0 & 0.0 & \textbf{3.3} & 0.0 & 0.3 & 0.8 \\
\arrayrulecolor{black!22}\midrule[0.3pt]\arrayrulecolor{black}
\rowcolor{vcbBestRowLight}\vcblogo{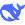}DeepSeek V4.1 Flash & max & 0.0 & 0.0 & 0.0 & 0.0 & 0.0 & 0.0 \\
\arrayrulecolor{black!22}\midrule[0.3pt]\arrayrulecolor{black}
\rowcolor{vcbBestRowLight}\vcblogo{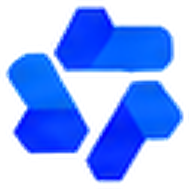}Qwen 3.8 Max & xhigh & 0.0 & 0.0 & 0.0 & 0.0 & 0.0 & 0.0 \\
\arrayrulecolor{black!22}\midrule[0.3pt]\arrayrulecolor{black}
\rowcolor{vcbBestRowLight}\vcblogo{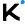}Kimi K3.1 Pro & max & 0.0 & 0.0 & 0.0 & 0.0 & 0.0 & 0.0 \\
\bottomrule
\end{tabularx}
\endgroup
\end{table*}

Table~\ref{tab:exp-32-t01} lists all 29 configurations, with models ordered by their best configuration; below we compare models by their best configurations. GPT-6 Astra (max) performs best, solving 10.7 of 40 tasks per run on average, an accuracy of 26.7\%; GPT-6.1 Sol (max) and Claude Opus 5.5 (xhigh) tie for second at 8.7/40 (21.7\%). No configuration reaches 30\%: even the best configuration leaves about three quarters of the tasks unsolved on average. In terms of category accuracy, the leading configurations score almost entirely on spatial manipulation and simulation: GPT-6 Astra (max) reaches 80.0\% there, whereas no configuration exceeds 13.3\% on any of the other three categories.

\medskip
The remaining five models lag far behind. GPT-6 Luna and Claude Sonnet 5.5 solve only 0.3 of 40 tasks per run in their best configurations, that is, a single correct answer across three runs, while DeepSeek V4.1 Flash, Qwen 3.8 Max, and Kimi K3.1 Pro solve no task in any of their three runs. The models therefore split sharply: the best configurations of the top three models solve 8.7 to 10.7 tasks per run, the other five solve almost none, and nothing lies in between. Visual Cipher Bench is thus far from saturated and separates the three leading models from the rest.

\medskip
\subsection{Analysis of Agent Performance}

\subsubsection{Rule Discovery Is the Main Bottleneck}
\label{sec:exp-bottleneck}

Pooling the 3,480 attempts of all 29 configurations by category, spatial manipulation and simulation reaches 22.8\% accuracy, whereas knowledge-based decoding, logical constraints and planning, and rule induction reach only 3.3\%, 2.4\%, and 0.6\%; the three leading models share the same profile (Table~\ref{tab:exp-32-t01}). In spatial tasks, the operations are usually given in the images, and the difficulty lies in executing a multi-step transformation correctly; rule-induction tasks instead require discovering a task-specific rule from a few examples. The latter are almost never solved: only 5 of 870 attempts are correct, and some of these trajectories show that the model never recovered the rule and guessed the answer only because its letters formed a theme-related word. Logic tasks are also low because they require large searches under constraints; Section~\ref{sec:exp-failure} shows that agents on these tasks often replace pruning with exhaustive enumeration.

\begin{table}[t]
\centering
\caption{Hint experiment: accuracy (\%) on the ten selected tasks. Columns: no hint, representation hint only, mechanism hint only, and both. The upper block reports each model; the lower block pools the three models by category. Each model runs each task three times under each condition.}
\label{tab:hint-ablation}
\begingroup
\small
\setlength{\tabcolsep}{8pt}
\setlength{\aboverulesep}{0pt}
\setlength{\belowrulesep}{0pt}
\renewcommand{\arraystretch}{1.3}
\begin{tabular}{lcccc}
\toprule
& None & Repr. & Mech. & Both \\
\midrule
\vcblogo{openai}GPT-6 Astra (max) & 23.3 & 23.3 & 70.0 & 76.7 \\
\vcblogo{openai}GPT-6.1 Sol (max) & 13.3 & 20.0 & 66.7 & 73.3 \\
\vcblogo{openai}GPT-6 Luna (max) & 0.0 & 6.7 & 16.7 & 23.3 \\
\arrayrulecolor{black!22}\midrule[0.3pt]\arrayrulecolor{black}
P1 Spatial & 38.9 & 44.4 & 61.1 & 66.7 \\
P2 Logic & 3.7 & 14.8 & 22.2 & 29.6 \\
P3 Rule & 0.0 & 3.7 & \textbf{85.2} & 88.9 \\
P4 Knowledge & 16.7 & 11.1 & 33.3 & 44.4 \\
\midrule
\rowcolor{vcbBestRowLight}Overall & 12.2 & 16.7 & 51.1 & 57.8 \\
\bottomrule
\end{tabular}
\endgroup
\end{table}

If rule discovery is indeed the main bottleneck, giving the rule directly should substantially raise accuracy. To test this, we design a hint experiment: we select ten tasks across the four categories and write for each a mechanism hint that states its core rule. A gain in accuracy, however, need not come from the rule itself; it could instead come from guidance on how to represent the task state, or from the added hint text itself. To rule out both explanations, we also write for each task a representation hint as a control: it is appended in the same way as the mechanism hint but only describes how the task state is represented, without stating the rule. Neither hint reveals the answer or any intermediate state (Appendix~\ref{app:hints}). We run GPT-6 Astra~(max), GPT-6.1 Sol~(max), and GPT-6 Luna~(max) under four conditions (no hint, representation hint only, mechanism hint only, and both), three times per task under each condition.

Table~\ref{tab:hint-ablation} reports the results. The mechanism hint raises overall accuracy from 12.2\% to 51.1\%, and both leading models rise from 13--23\% to about 70\%; the largest gain is in rule induction (0.0\%$\rightarrow$85.2\%). Under the same time limit, timeouts also fall from 51 to 29. By contrast, the representation hint alone yields only a limited gain (16.7\%), indicating that the improvement comes from the rule itself rather than from representation guidance or the hint text. Once the rule is known, the leading models can thus carry out the remaining multi-step reasoning in most attempts. Even with the mechanism given, however, logic tasks reach only 22.2\%, and the weaker GPT-6 Luna rises only to 16.7\%, indicating that in these two cases execution itself remains an obstacle.

\subsubsection{Failure Modes}
\label{sec:exp-failure}

Unsolved attempts end in one of two main ways: the agent submits a wrong answer, or it does not finish within the 60-minute limit. We manually audit representative wrong-answer and timeout trajectories and analyse their causes against the puzzle, the walkthrough, and the reference solver.

\paragraph{Wrong answers: insufficient verification.}
Most wrong answers are not computational slips: the agent does not test its hypothesis against evidence already on the page. Most often, once it finds a rule that explains some of the examples, it answers without checking the rule against the remaining examples or the global constraints. When the rule stays unresolved, agents also tend to abandon induction and submit an answer that fits the puzzle's theme but is not supported by the examples, sometimes even while acknowledging that the answer is incomplete. By contrast, trajectories that find the correct mechanism but err in transcription or execution are less common. Successful trajectories do the opposite: they write several candidate rules, turn every example into an assertion to filter them, and only then work out the targets. The examples in a puzzle are themselves test cases, and agents that answer wrongly rarely make full use of them.

\paragraph{Timeouts: enumeration in place of constraints.}
Trajectories that time out do not lack tool use; the problem is that they spend their compute on enumeration. Some search broadly through generic rule families such as cellular automata and morphological operators, or keep switching among interpretations; these candidates do not come from cues in the images, and the agents write no test that quickly rules them out, so the search space never contracts. Others find the correct mechanism but try every combination of unknown states instead of pruning with the puzzle's constraints, and cannot finish in time. The hint experiment confirms this from the other side: once the mechanism hint narrows the hypothesis space, timeouts fall sharply under the same time limit. Most timeouts therefore reflect not a lack of time but a hypothesis space that the visual evidence never narrowed.

Both failure modes thus point to the same deficiency: agents rarely use visual evidence to constrain their hypotheses. Wrong answers use only part of the evidence; timeouts rely on enumeration rather than evidence to narrow the search.

\subsection{Benchmark Validation}

\begin{figure*}[t]
\centering
\input{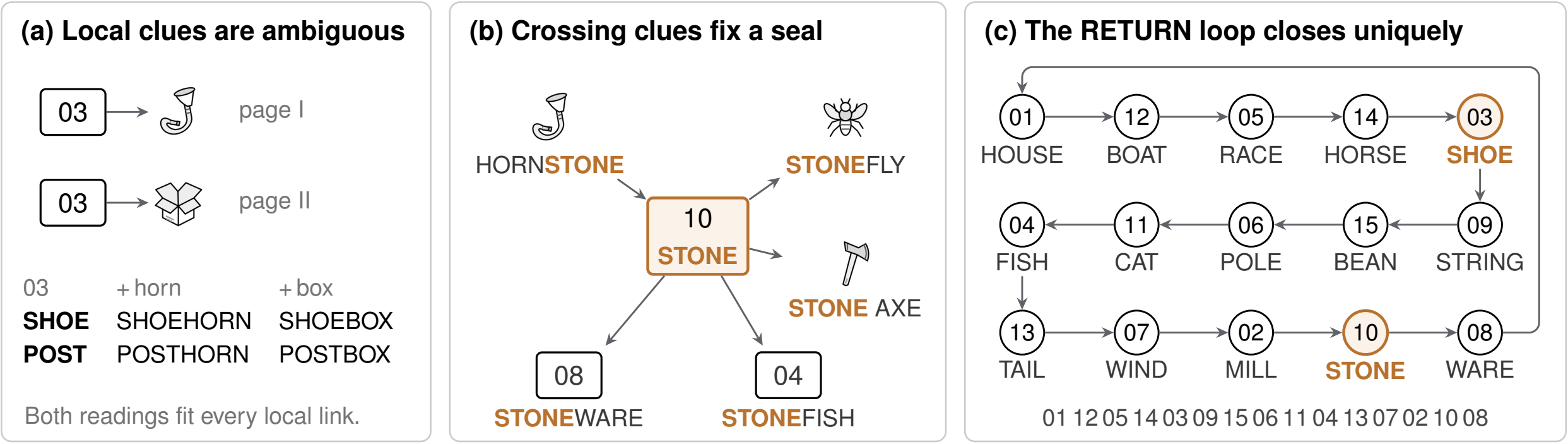}
\caption{Solving Closed Stock (knowledge-based decoding). \textbf{(a)} Seal 03 points to a horn on page I and to a box on page II; both SHOE and POST satisfy these two local clues. \textbf{(b)} Seal 10 takes part in five compounds, and only STONE satisfies all of them, fixing seals 10, 04, and 08. \textbf{(c)} RETURN requires the 15 words to form a closed loop of compounds; only one such loop exists, and it also rules out POST. Reading the seal numbers from 01 gives the answer. Icons are taken from the puzzle; the full puzzle is shown in Appendix~\ref{app:case-puzzles}.}
\label{fig:exp-343-cases}
\end{figure*}

\subsubsection{Reliability and Solvability}

Grading in Visual Cipher Bench is deterministic exact matching: the same submission always receives the same score, so randomness in the evaluation comes only from the agent's solving process. We therefore run each model--effort configuration independently three times on all 40 tasks and report the mean accuracy over the three runs, in line with recommendations to account for sampling variability in model evaluations~\citep{miller2024errorbars}.

\medskip
Every task comes with a step-by-step walkthrough and a reference solver. The walkthrough records the visual evidence, rule, and disambiguation behind each key step, and the reference solver reproduces the reference answer from the puzzle images alone. An oracle agent that runs the reference solver passes the grader on all 40 tasks. In addition, two human experts reviewed every task and its walkthrough, confirming that the visual evidence suffices to derive the answer, that every step of the walkthrough is sound, and that the answer is unique; all 40 tasks passed the review.

\medskip
\subsubsection{Case Study: Closed Stock}
\label{sec:vcb-cases}

We use Closed Stock, a knowledge-based decoding task (Figure~\ref{fig:exp-343-cases}; full puzzle in Appendix~\ref{app:case-puzzles}), to show how a puzzle satisfies both design principles, solvability and long-horizon reasoning; this category reaches only 3.3\% accuracy across all configurations. The puzzle is a two-page register in which 15 numbered seals and several pictured objects are linked by arrows. The same seal number appears repeatedly across both pages, yet the register gives each number a single blank, so each number must take one word; at the bottom, a RETURN chain of 15 empty circles winds through three rows and back to 01. The solver must infer the mechanism unaided: two items joined by an arrow form a common English compound, and the 15 words must also chain into a closed loop.

\medskip
Solving proceeds through three layers of constraints. First, local clues are ambiguous (Figure~\ref{fig:exp-343-cases}a): seal 03 points to a horn on page I and to a box on page II, and both SHOE (SHOEHORN, SHOEBOX) and POST (POSTHORN, POSTBOX) satisfy the two clues. Second, crossing clues fix some seals (Figure~\ref{fig:exp-343-cases}b): seal 10 is linked to the horn, a fly, an axe, and seals 04 and 08, and only STONE satisfies all five compounds HORNSTONE, STONEFLY, STONE AXE, STONEFISH, and STONEWARE, which fixes seals 10, 04, and 08 as STONE, FISH, and WARE; the constraints then propagate across pages, for instance WAREHOUSE fixes seal 01 as HOUSE. Finally, the global loop yields a unique answer (Figure~\ref{fig:exp-343-cases}c): several seals still admit multiple candidates at this point (seal 02 may be MILL or WATER, seal 07 WIND, AIR, or MAIL, and seal 03 is still SHOE or POST), but only one loop links all 15 words end to end, HOUSE–\allowbreak BOAT–\allowbreak RACE–\allowbreak HORSE–\allowbreak SHOE–\allowbreak STRING–\allowbreak BEAN–\allowbreak POLE–\allowbreak CAT–\allowbreak FISH–\allowbreak TAIL–\allowbreak WIND–\allowbreak MILL–\allowbreak STONE–\allowbreak WARE, and its links HORSESHOE and SHOESTRING settle seal 03 as SHOE. Reading the seal numbers from 01 gives the answer \texttt{011205140309150611041307021008}.

\medskip
The puzzle embodies both design principles. It is solvable: every inference is supported by the puzzle page, less common compounds such as HORNSTONE and MILLRACE are each cross-confirmed by several familiar ones, and the final answer is unique. It also requires long-horizon reasoning: no single clue reveals the answer, and the solver must propagate constraints across two pages, hold ambiguities open, and finally check every choice against the global loop. Agent behaviour bears this out. The successful Claude Opus 5.5 trajectory keeps local ambiguities open and uses the complete loop to fix the order, matching the reference solution.

\section{Related Work}
\label{sec:related-work}

\noindent\textbf{Visual reasoning benchmarks.}
Most visual reasoning benchmarks evaluate image understanding through a given question. MMMU~\citep{yue2024mmmu} and MMMU-Pro~\citep{yue2025mmmupro} require models to interpret charts and diagrams in exam questions using subject knowledge, while MathVista~\citep{lu2024mathvista} and MathVerse~\citep{zhang2024mathverse} test whether models can extract quantitative and geometric relations from images to answer mathematical questions. Perception probes show that multimodal models still err on judgments that humans make at a glance~\citep{fu2024blink,tong2024eyes,rahmanzadehgervi2024blind}. For spatial simulation, mental-rotation experiments show that humans mentally simulate transformations of figures~\citep{shepard1971mental}, and STARE~\citep{li2026stare} evaluates multi-step visual simulation in tasks such as folding cube nets. Rule induction forms another line: Raven's Progressive Matrices~\citep{carpenter1990raven} and their procedurally generated counterpart RAVEN~\citep{zhang2019raven} ask solvers to complete a pattern matrix, while ARC~\citep{chollet2019measure} and ARC-AGI-2~\citep{chollet2025arcagi2} require inferring grid transformations from a few input--output examples. Models can propose natural-language hypotheses and implement them as programs for testing~\citep{wang2024hypothesis}, yet high accuracy does not imply that the intended abstraction has been acquired~\citep{beger2025competence}, and part of the difficulty on visual inputs stems from perception rather than induction~\citep{wang2025perception}. In these benchmarks, the question or task format specifies the task, and the reasoning chains are short. Visual Cipher Bench provides no such framing: the agent must determine from the images which graphics are examples and which mechanism to apply.

\medskip
\noindent\textbf{Long-horizon agent benchmarks.}
As reasoning, acting, and tool use have improved~\citep{wei2022cot,yao2023react,schick2023toolformer,wang2024codeact,shinn2023reflexion}, language-model agents have become increasingly capable of long-horizon problem solving. The corresponding evaluations focus mainly on text and code tasks: SWE-bench~\citep{jimenez2024swebench} and SWE-Bench Pro~\citep{deng2025swebenchpro} require resolving issues in real code repositories, SWE Refactor Bench~\citep{hong2026swerefactorbench} requires migrating the technology stack of an entire repository, Terminal-Bench~\citep{merrill2026terminalbench} requires completing complex command-line tasks, and MLE-bench~\citep{chan2025mlebench}, WebArena~\citep{zhou2024webarena}, and GAIA~\citep{mialon2024gaia} cover machine-learning engineering, web navigation, and general assistant tasks, respectively, while Frontier-Eng~\citep{chi2026frontiereng} and AI4AI-Bench~\citep{chi2026ai4aibench} evaluate the iterative optimization of engineering designs and the design of training algorithms. Meanwhile, the length of tasks that agents can complete autonomously is growing rapidly~\citep{kwa2025measuring}. Work on the visual side, by contrast, rarely probes long-horizon reasoning. Visual programming, Visual Sketchpad, and ``thinking with images'' approaches allow models to crop, annotate, or transform images while reasoning~\citep{gupta2023visprog,suris2023vipergpt,hu2024visualsketchpad,zheng2025deepeyes,su2025thinking}, but they are mainly validated on single-turn visual question answering and yield limited gains over direct reasoning~\citep{wang2026illusion}. In graphical-interface tasks such as OSWorld~\citep{xie2024osworld}, images mainly serve localization and operation; in 3D creation tasks such as Frontier3DBench~\citep{lin2026frontier3dbench}, they mainly serve as references for reconstruction and generation. Interactive games such as BALROG, LongPuzzleBench, and ARC-AGI-3 require many actions, but the environment returns a new state after every action, so agents can obtain feedback by trial and error rather than reasoning solely from evidence in fixed images~\citep{paglieri2025balrog,zhang2026longpuzzlebench,arcprize2026arcagi3}. Long-horizon reasoning over fixed visual evidence therefore still lacks a dedicated benchmark.

\medskip
\noindent\textbf{Puzzles and cipher solving.}
Puzzles have long been used to study human problem solving: cognitive science has found that breakthroughs often come from restructuring the problem representation rather than from searching harder within it~\citep{duncker1945problem,newell1972human,kaplan1990insight}. Puzzles are now also used to evaluate language models: cryptic crosswords~\citep{rozner2021cryptic}, rebuses~\citep{gritsevskiy2024rebus,lee2025puzzled}, and visual riddles~\citep{bittonguetta2024visualriddles} require combining wordplay, graphical layout, and commonsense; shift-cipher experiments show that decoding accuracy varies with how common a shift is in the training data~\citep{mccoy2024embers}; and multimodal puzzle collections cover abstract figures, algorithmic puzzles, and IQ-test-style problems~\citep{chia2024puzzlevqa,ghosal2025algopuzzlevqa,cai2025mmiq,song2026visualpuzzles,lymperaiou2026probing,kuang2025express}. EnigmaEval~\citep{wang2025enigmaeval} and PuzzleWorld~\citep{li2026puzzleworld} further use puzzle-competition problems to evaluate open-ended reasoning. However, their puzzles are drawn from existing human puzzle events and mix textual and visual clues, and as models have improved, most of these problems have already been solved. In contrast, all puzzles in Visual Cipher Bench are original, and their rules, examples, and clues exist only in images. Each puzzle is designed under the principles of solvability and long-horizon reasoning and reviewed by human experts to ensure both quality and difficulty; the best configuration currently solves only about a quarter of the puzzles.

\section{Conclusion}

We presented Visual Cipher Bench: 40 original visual puzzles whose rules, examples, and clues all reside in one to three black-and-white line drawings, which agents must solve over an extended period with tools in an offline sandbox and which are scored only by exact match on the final answer. Across 29 reasoning-effort configurations of eight frontier models, the best configuration, GPT-6 Astra (max), solves only 10.7 of 40 tasks on average (26.7\%), and five of the eight models solve fewer than one task on average in their best configurations; accuracy reaches 22.8\% on spatial manipulation and simulation but only 0.6\% on rule induction. On a ten-task subset covering all four categories, a single mechanism hint that does not reveal the answer raises accuracy from 12.2\% to 51.1\%, indicating that the main weakness of current agents lies not in executing multi-step plans but in discovering the rules from visual evidence on their own. Improving long-horizon visual reasoning therefore hinges on teaching agents to discover rules from images and keep checking them; Visual Cipher Bench provides a reliable benchmark for measuring this capability.

\par\FloatBarrier

\bibliographystyle{unsrtnat}
\begingroup
\renewcommand{\bibfont}{\fontsize{8.5}{10}\selectfont}
\setlength{\bibsep}{1pt plus 0.5pt}
\bibliography{ref,related_refs}
\endgroup

\clearpage
\appendix
\section{Evaluation Setup}
\label{app:eval-setup}

This appendix details the evaluation described in Section~\ref{sec:vcb-evaluation}: the submission instruction shared by all tasks is shown below, and the runtime environment and grading details are listed in Table~\ref{tab:eval-setup}.

\begin{tcolorbox}[enhanced,colback=black!3,colframe=black!3,boxrule=0pt,sharp corners,
  borderline west={1.4pt}{0pt}{black!75},left=4mm,right=4mm,top=2mm,bottom=2mm,
  before skip=8pt,after skip=10pt]
{\sffamily\bfseries\footnotesize Submission instruction (shared by all 40 tasks)}\par\smallskip
\small Solve this visual reasoning puzzle: examine all images in \texttt{/app/images/} and write the single final answer to \texttt{/work/answer.txt}. Write only the answer itself, with no quotes, punctuation, or explanation.
\end{tcolorbox}

\begin{table}[h]
\centering
\caption{Runtime environment and grading settings.}
\label{tab:eval-setup}
\small
\renewcommand{\arraystretch}{1.15}
\begin{tabularx}{\linewidth}{@{}lX@{}}
\toprule
Item & Setting \\
\midrule
Harness & Harbor v0.24.0~\citep{shi2026harbor} \\
Agent runner & Codex CLI v0.163.0 for GPT-family models; Claude Code CLI v2.1.296 for all other models \\
Time limit & 60 minutes per task attempt \\
Compute & 2 CPUs, 4\,GB memory, no GPU \\
Environment & Network-disabled container with Python 3.12, NumPy, and Pillow; puzzle images are read-only \\
Runs & 40 tasks \(\times\) 3 runs \(=\) 120 attempts per model--effort configuration \\
Answer normalization & Unicode NFKC normalization, whitespace removal, and uppercasing; for alphanumeric answers, commas, hyphens, or arrows may separate elements (e.g., \texttt{5,4,2} and \texttt{542} are equivalent) \\
Invalid submissions & A missing answer, a symbolic link at the answer path, or a file larger than 4\,KB scores 0 \\
\bottomrule
\end{tabularx}
\end{table}

\section{Case-Study Puzzle}
\label{app:case-puzzles}

This appendix shows the complete Closed Stock puzzle analysed in Section~\ref{sec:vcb-cases}, i.e., the two original images given to the evaluated agents.

\begin{figure}[H]
\centering
\includegraphics[width=0.45\linewidth]{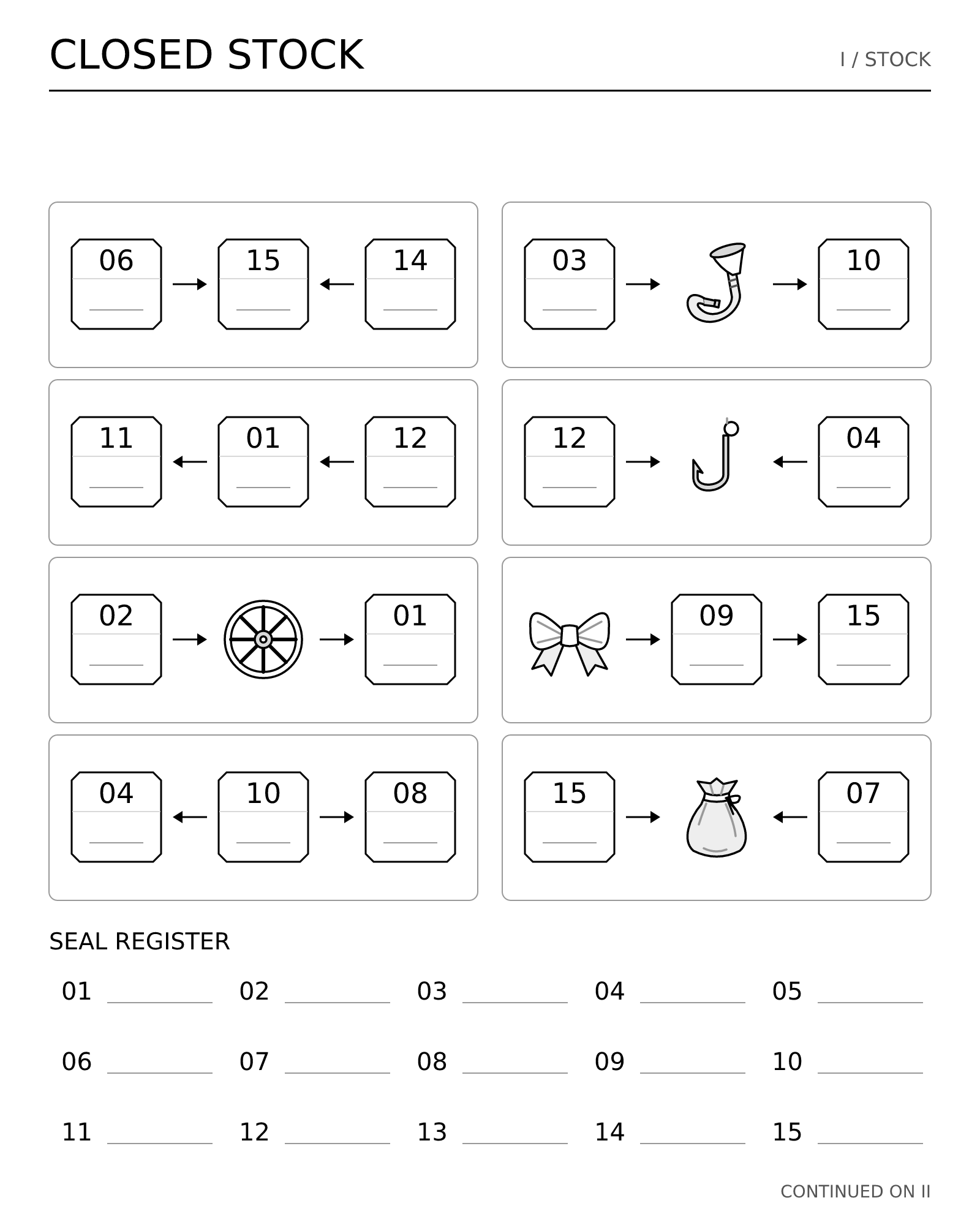}\hspace{0.04\linewidth}%
\includegraphics[width=0.45\linewidth]{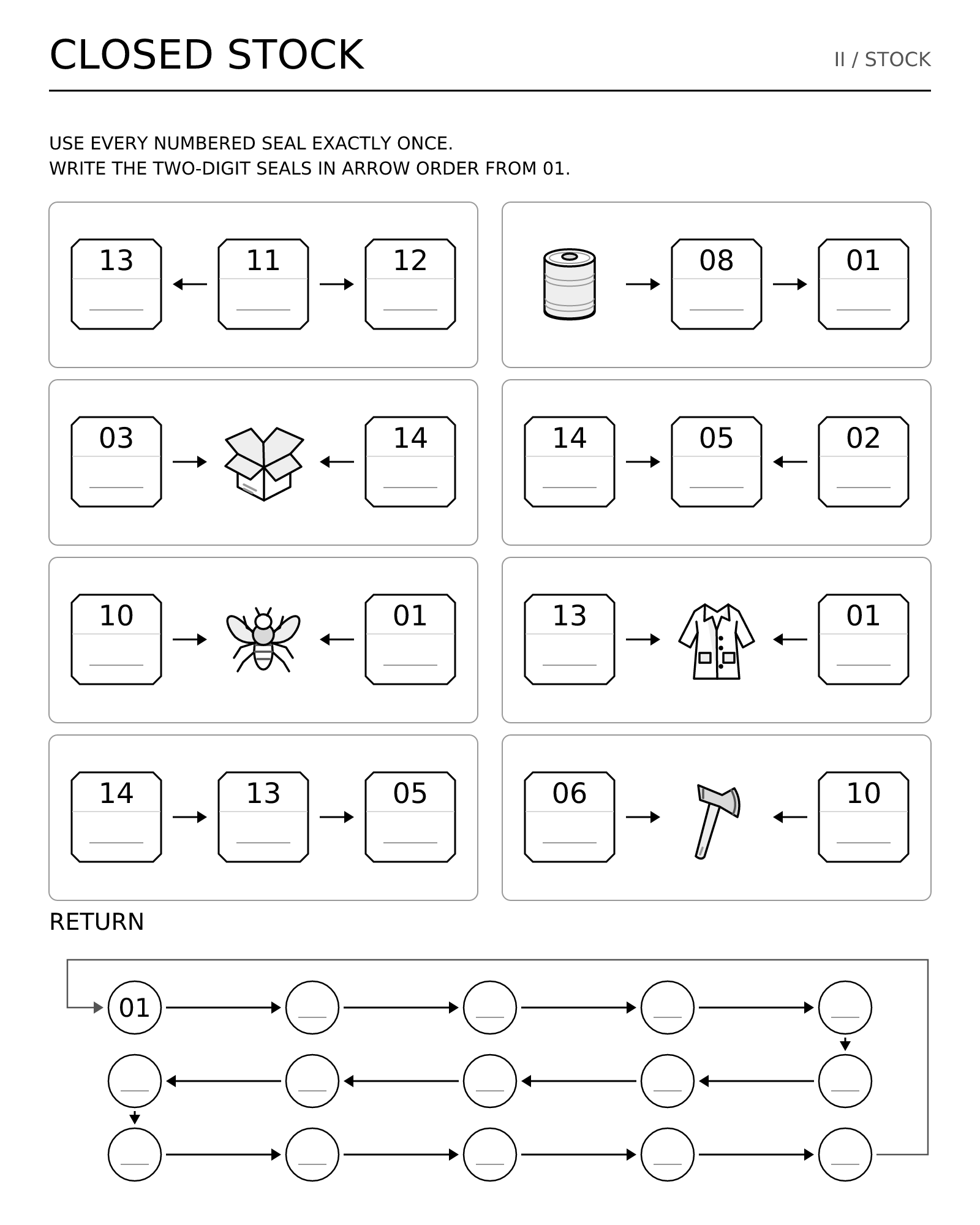}
\caption{The complete Closed Stock puzzle.}
\label{fig:app-closed-stock}
\end{figure}

\section{Hint Experiment}
\label{app:hints}

This appendix lists the hints used in the hint experiment of Section~\ref{sec:exp-bottleneck}.

\begin{table}[h]
\centering
\caption{Representation and mechanism hints, grouped by category.}
\label{tab:app-hints}
\footnotesize
\renewcommand{\arraystretch}{1.2}
\begin{tabularx}{\linewidth}{@{}llXX@{}}
\toprule
& Task & Representation hint & Mechanism hint \\
\midrule
P1 & \texttt{the-shutter-loft} & Treat the blades as opaque polygons stacked in an unknown bottom-to-top layer order, each carrying a dot that moves with it, and treat the checkpoints as complete snapshots of that stack. & In each round the blade with the largest total exposed area departs, the counted blade turns a quarter turn clockwise about the departing blade's dot and is raised to the top, and the blade immediately beneath then turns a quarter turn anticlockwise about the raised blade's new dot. \\
P1 & \texttt{window-dressing} & The two pages each show eight open frames of the same eight letters under two different views, and each letter's strokes are split across the two views, so a frame shows only part of what belongs to it. & The letters are ordered by how far each moves horizontally between the two views, from largest displacement to smallest, and each letter is the one whose strokes fit both views without placing black in any exposed white region. \\
\arrayrulecolor{black!22}\midrule[0.3pt]\arrayrulecolor{black}
P2 & \texttt{echo-vault} & Each specimen sits in a numbered square and also carries its own circular ring value; the two inventories are separate, and the labelled columns are complete snapshots rather than a list of steps. & For each named pair, look up each word's current ring to find which snapshot gives its square, exchange the two tubes at those squares carrying their rings along, and then exchange only the rings at the two slots named by the original current ring values. \\
P2 & \texttt{jacquard-memory} & The board is split by one fixed closed loop into inside and outside cells, initial values are hidden, and the loop does not travel with the letters. & A move cyclically shifts a whole row or column by the sum of the values currently in that line's inside cells, and a value that crosses the loop boundary becomes four minus itself. \\
P2 & \texttt{latent-exposure} & The grid is divided by a fixed closed loop into inside and outside positions, the hidden numbers are not printed, and the loop does not travel with the letters. & Each move shifts a whole row right or a whole column down by the sum of the current numbers in that line's inside positions, and a number carried across the loop becomes four minus itself while its letter is unchanged. \\
\arrayrulecolor{black!22}\midrule[0.3pt]\arrayrulecolor{black}
P3 & \texttt{dockside-detours} & Counters sit at fixed numbered positions on a ring and stay there; only the letters move between those positions, and the circle marks the current position. & Each round decrements the current counter, moves every letter whose counter now equals that new value forward to the next such position and wraps, then hops forward from the departing letter's new position by the number of positions whose visible letter changed; stop when the current counter reaches zero. \\
P3 & \texttt{echo-basin} & Read each dial as a lattice of sample positions whose states are empty, diamond or ring, so that a contact transforms the incoming pattern into a new pattern and a returned control. & The difference between the incoming and outgoing pattern is a clockwise rotation of the incoming pattern by one quarter turn for each ring it contains. The returned control then moves its rows right by counts taken from the guide pattern after its rows have already moved, and its columns down by counts taken from that same guide pattern. \\
P3 & \texttt{fickle-neighbors} & Each cell holds a card with its own arrow and shade, a card may carry no letter, and the visitor is marked by the circle and moves from cell to cell. & The source card's arrow is saved, the visitor moves one square along it, the departure card flips shade, the whole source row cycles right if it was white or left if grey, and every arrow in the source column then rotates by the angle that makes the incoming card point along the saved direction. \\
\arrayrulecolor{black!22}\midrule[0.3pt]\arrayrulecolor{black}
P4 & \texttt{cutting-corners} & Every white run in the grid is a word, a picture cell stands for one whole word shared by the run across and the run down, and the stencils supply letters into empty cells. & Each stencil is turned but never flipped and laid so that its letters fill empty cells, and each picture is read as the sound that completes both its across and its down word. \\
P4 & \texttt{firm-friends} & The pictured objects form six reciprocal pairs, and each pair yields two words of fixed length that must fit the across or down entry of their own picture. & Two paired names exchange their initial consonant sounds and are respelled, and each result enters the entry belonging to the ending of the picture it came from. \\
\bottomrule
\end{tabularx}
\end{table}

\end{document}